\documentclass[letterpaper]{article} % For LaTeX2e
\usepackage{iclr2023_conference,times}

\usepackage{amsmath,amsfonts,bm}

\def\Figref#1{Figure~\ref{#1}}

\def\eqref#1{equation~\ref{#1}}
\def\1{\bm{1}}

\DeclareMathAlphabet{\mathsfit}{\encodingdefault}{\sfdefault}{m}{sl}
\SetMathAlphabet{\mathsfit}{bold}{\encodingdefault}{\sfdefault}{bx}{n}

\usepackage{hyperref}
\usepackage{url}
\usepackage[utf8]{inputenc} % allow utf-8 input
\usepackage[T1]{fontenc}    % use 8-bit T1 fonts
\usepackage{hyperref}       % hyperlinks
\usepackage{url}            % simple URL typesetting
\usepackage{booktabs}       % professional-quality tables
\usepackage{amsfonts}       % blackboard math symbols
\usepackage{nicefrac}       % compact symbols for 1/2, etc.
\usepackage{microtype}      % microtypography
\usepackage{xcolor}         % colors
\usepackage{amsmath}
\usepackage{bbding}
\usepackage{graphicx} %插入图片的宏包
\usepackage{float} %设置图片浮动位置的宏包
\usepackage{subfigure} %插入多图时用子图显示的宏包
\usepackage{xspace}
\definecolor{jscolornotes}{rgb}{0.2,0.6,0.2}
\definecolor{sicolor}{rgb}{0.1,0.1,0.6}

\newcommand{\jsremove}[1]{} % to remove JSnotes
\definecolor{xxawfcolor}{rgb}{0.6,0.2,0.2}

\newcommand{\IGNORE}[1]{{}}

\newcommand{\be}{\begin{equation}}
\newcommand{\ee}{\end{equation}}
\newcommand{\bea}{\begin{eqnarray}}
\newcommand{\eea}{\end{eqnarray}}
\newcommand{\beas}{\begin{eqnarray*}}
\newcommand{\eeas}{\end{eqnarray*}}

\newcommand{\cut}[1]{}

\newcommand{\pixel}{x}
\newcommand{\vertex}{v}
\newcommand{\vertices}{\mathcal{V}}
\newcommand{\mesh}{\mathcal{M}}
\newcommand{\allstages}{S}
\newcommand{\stage}{s}
\newcommand{\downsampledfeature}{F}
\newcommand{\upsampledfeature}{H}
\newcommand{\imagefeature}{Q}
\newcommand{\meshfeature}{G}

\newcommand{\allfaces}{\mathcal{T}}
\newcommand{\face}{t}
\newcommand{\seq}{l}
\newcommand{\edge}{e}
\newcommand{\facenormal}{n}
\newcommand{\poseheatmap}{P}
\newcommand{\gcnkernel}{g}
\newcommand{\seqorder}{O}
\newcommand{\silhouetteheatmap}{C}
\newcommand{\poseheatmappixel}{x^\poseheatmap}
\newcommand{\silhouetteheatmappixel}{x^\silhouetteheatmap}
\newcommand{\seqlength}{L}
\newcommand{\ie}{{\emph{i.e.}},~}

\newcommand{\eg}{{\emph{e.g.}},~}

\iclrfinalcopy
\title{Pixel-Aligned Non-parametric Hand Mesh Reconstruction}
\author{Shijian Jiang$ ^1$, Guwen Han$ ^1$, Danhang Tang$ ^2$, Yang Zhou$ ^3$, Xiang Li$ ^3$, Jiming Chen$ ^1$, Qi Ye$ ^1$  \\
\thanks{$^1$ Zhejiang University, $^2$ Google, $^3$ OPPO, Emails:\texttt{\{jsj630, qi.ye\}@zju.edu.cn}} \\
}
\begin{document}

\maketitle

\begin{abstract}
Non-parametric mesh reconstruction has recently shown significant progress in 3D hand and body applications. In these methods, mesh vertices and edges are visible to neural networks, enabling the possibility to establish a direct mapping between 2D image pixels and 3D mesh vertices. In this paper, we seek to establish and exploit this mapping with a simple and compact architecture. The network is designed with these considerations: 1) aggregating both local 2D image features from the encoder and 3D geometric features captured in the mesh decoder; 2) decoding coarse-to-fine meshes along the decoding layers to make the best use of the hierarchical multi-scale information.  Specifically, we propose an end-to-end pipeline for hand mesh recovery tasks which consists of three phases: a 2D feature extractor constructing multi-scale feature maps, a feature mapping module transforming local 2D image features to 3D vertex features via 3D-to-2D projection, and a mesh decoder combining the graph convolution and self-attention to reconstruct mesh. The decoder aggregate both local image features in pixels and geometric features in vertices. It also regresses the mesh vertices in a coarse-to-fine manner, which can leverage multi-scale information. By exploiting the local connection and designing the mesh decoder, Our approach achieves state-of-the-art for hand mesh reconstruction on the public FreiHAND dataset.
\end{abstract}

\section{Introduction}
\par Reconstructing 3D hand mesh from a single RGB image has attracted tremendous attention as it has numerous applications in human-computer interactions (HCI), VR/AR, robotics, \emph{etc}. Recent studies have made great efforts in the accurate hand mesh reconstruction and achieved very promising results~\citep{metro,meshnet,weakly,ge,hasson}. Recent state-of-the-art approaches address the problem mainly by deep learning. These learning-based methods can be roughly divided into two categories according to the representation of the hand meshes, \ie the parametric approaches, and the non-parametric ones. The parametric approaches use a parametric model that projects hand meshes in a low dimensional space (\eg MANO~\citep{mano}) and regresses the coefficients in the space (\eg the shape and pose parameters of MANO) to recover the 3D hand~\citep{hasson}. The non-parametric ones instead directly regress the mesh vertices using graph convolution neural network~\citep{meshnet,weakly,handmesh} or transformer~\citep{metro}.
\par Non-parametric approaches have shown substantial improvement over the parametric ones in recent work, owing to the mapping between the image and the vertices is less non-linear than that between the image and the coefficients of the hand models~\citep{goal}. Their pipelines ~\citep{weakly,handmesh,metro} usually consist of three stages: a 2D encoder extracts the global image feature, which is mapped to 3D mesh vertices before fed into a 3D mesh decoder operating on the vertices and edges to get the final mesh. 
\par Despite the success, the potential of non-parametric approaches has not been fully uncovered with this pipeline. In parametric methods, vertices and edges are not visible to the network, and no operation is carried out in the manifold of the meshes; 2D image features are extracted only to learn a mapping between the image content and the hand model parameters.
Conversely in non-parametric methods, operations on vertices and edges, such as graph convolutions or attention modules, are designed to aggregate the geometric features of the meshes. With the operation, vertices and edges are visible to the networks; thus direct connections between pixels of the 2D image feature space and vertices of the 3D mesh can be established and operations in the decoder can aggregate both image features and geometric features, which can not be realized in the parametric methods. This connection and the aggregation, however, have not been fully explored by previous work.
\par In this paper, we seek to establish the connections and merge the local hand features from appearance in the input and geometry in the output. To this end, we utilize the pixel-aligned mapping module to establish the connections and propose a simple and compact architecture to deploy the connections. We design our network by making the following philosophical choices: 1) For the 2D feature extractor, we keep feature maps of different scales in the encoder instead of using the final global feature to enable 2D local information mapping to 3D. 2) We decode the mesh in the coarse-to-fine manner to make the best use of the multi-scale information. 3) Both image features and geometric features are aggregated in the operations of the mesh decoder rather than only geometric features. 
% \yq{may need to be specific as describing the methods instead of these abstract concept. Is only geometric features correct? geometric feature + global feature is geometric feature only?}

% Our design is shown in Fig.\ref{fig_pipeline}. Multi-scale information of pixels in the image encoder is naturally passed to the vertices in the 3D mesh decoder. Our experiments show the design enables better alignment between the image and the reconstructed mesh. The aggregation of features not only improve the graph convolution network substantially but also gains large superiority over the attention mechanism with global features. \yq{to be refined}
\par
\begin{figure}[!htbp]
  \centering
  \vspace{-0.5cm}
  \includegraphics[width=\textwidth,trim={0 1cm 0 0cm},clip]{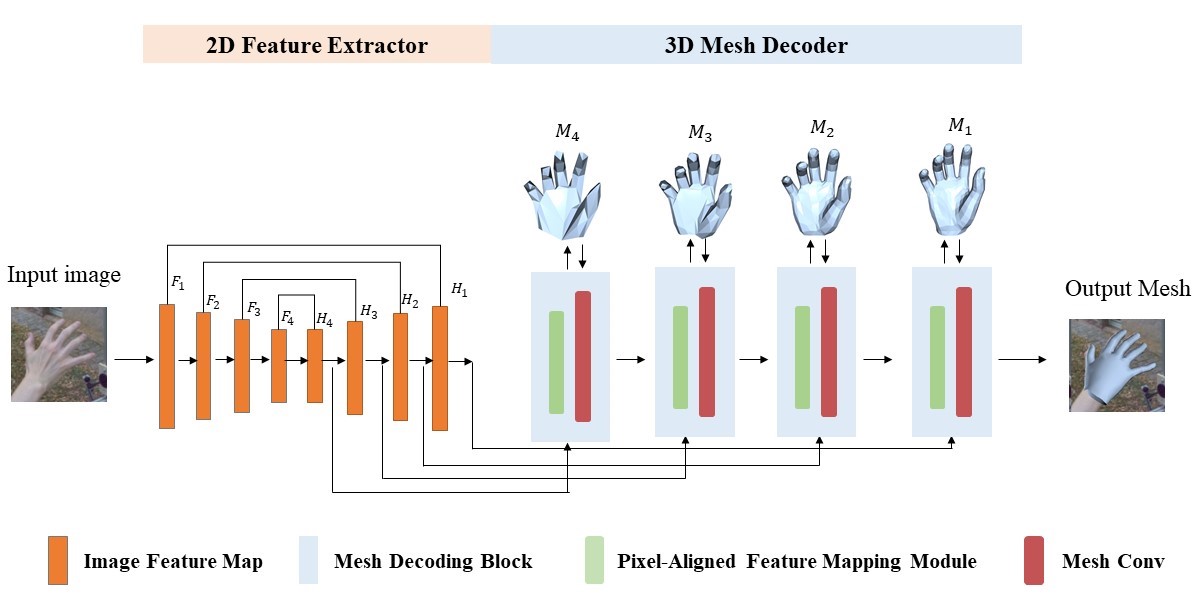}
  \caption{\small \textbf{Our full pipeline.} Given a single-hand RGB image as input, our network first extracts 2D feature maps with an hourglass module. Downsampled $\downsampledfeature_\stage$ and upsampled feature maps $\upsampledfeature_\stage$ are aggregated to have the multiscale image feature pyramids, which are mapped to vertices of meshes at different scales by our pixel-aligned mapping module and GCN blocks. Vertex position of the meshes can then be predicted in a coarse-to-fine manner.}
  \label{fig_pipeline}
\end{figure}

 Our design is shown in \Figref{fig_pipeline}. Multi-scale image features are naturally passed to the 3D mesh decoder. Our experiments show the design enables better alignment between the image and the reconstructed mesh. The aggregation of features not only improves the graph convolution network substantially but also gains large superiority over the attention mechanism with global features.

\par To summarize, our key contributions are 1) Operations are capable of aggregating both local 2D image features and 3D geometric features on meshes in different scales. 2) Connections between pixels of 2D image appearance in the encoder and vertices of 3D meshed in the decoder are established by a pixel-vertex mapping module. 3) A novel graph convolution architecture achieves state-of-the-art results on the FreiHAND benchmark.
%  \yq{Haven't got the results to support this}

\section{Related Work}
\paragraph{Mesh Reconstruction.} Previous research methods employ pre-trained parametric human hand and human models, namely MANO~\citep{mano}, SMPL~\citep{smpl}. And estimate the pose and shape coefficients of the parametric model. However, it is challenging to regress pose and shape coefficients directly from input images. Researchers propose to train network models with human priors, such as using skeletons~\citep{unitepeople} or segmentation maps. Some researchers have proposed regressed SMPL parameters by relying on human key points and contour maps~\citep{pavlakos2018learning,tan2017indirect} of the body. Coincidentally~\citep{omran2018neural} utilized the segmentation map of the human body as a supervision condition. A weakly supervised approach~\citep{kanazawa2018end} using 2D keypoint reprojection and adversarial learning regression SMPL parameters. Hsiao-Yu Tung~\citep{tung2017self} proposed a self-supervised approach to regression of human parametric models.
\par Recently, model-free methods~\citep{pose2mesh,meshnet,cmr} for directly regressing human pose and shape from input images have received increasing attention. Because it can express the nonlinear relationship between the image and the predicted 3D space. Researchers have explored various ways to represent the human body and hand using 3D mesh~\citep{metro,cmr,pose2mesh,graphformer,litany2018deformable,ranjan2018generating,verma2018feastnet,pixel2mesh,meshnet}, voxel spaces~\citep{bodynet}, or occupancy fields~\citep{pifu,niemeyer2020differentiable,xu2019disn,saito2020pifuhd,peng2020convolutional}. Among them, the voxel space method adopts a completely non-parametric method, which requires a lot of computing resources, and the output voxel needs to fit the body model to obtain the final human 3D mesh.Among the recent research methods, Graph Convolution Neural Networks (GCNs)~\citep{cmr,pose2mesh,graphformer,litany2018deformable,ranjan2018generating,verma2018feastnet,pixel2mesh,meshnet} is one of the most popular methods. Because GCN is particularly convenient for convolution operations on mesh data. However, GCN is good for representing the local features of the mesh, and the global features of the long-distance interaction between human vertices and joints cannot be well represented. Transformer-based methods~\citep{metro} use a self-attention mechanism to take full advantage of the information interaction between vertex and joints and use the global information of the human body to reconstruct more accurate vertex positions. But whether it is a GCN-based method or an attention mechanism-based method. Neither considers pixel-level semantic feature alignment information. Local pixel-level semantic feature alignment can compensate for the global information that GCN and transformer methods focus on.
\paragraph{Graph Neural Networks.} Graph deep learning generalizes neural networks to non-Euclidean domains, and we hope to apply graph convolution neural networks to learn shape-invariant features on triangular meshes. For example, spectral graph convolution neural network methods~\citep{bruna2013spectral,defferrard2016convolutional,kipf2016semi,levie2018cayleynets} perform convolution operations in the frequency domain. Local graph methods~\citep{masci2015geodesic,boscaini2016learning,monti2017geometric}  based on spatial graph convolutions make deep learning on Manifold more convenient.
\par In the application of mesh reconstruction.~\citep{ranjan2018generating} used fast local spectral filters to learn nonlinear representations of human faces.~\citep{kulon2019single} extended autoencoder networks to 3D representations of hands.Kolotouros proposed GraphCMR~\citep{cmr} to regression 3D mesh vertices using a GCN~\citep{cmr,pose2mesh,graphformer,litany2018deformable,ranjan2018generating,verma2018feastnet,pixel2mesh,meshnet}. Pose2Mesh~\citep{pose2mesh} proposes to reconstruct a human mesh from a given human pose representation based on a cascaded GCN.~\citep{simple} proposed spiral convolution to handle mesh in the spatial domain. Based on SpiralConv, Kulon ~\citep{weakly} introduced an automatic method to generate training data from unannotated images for 3D hand reconstruction and pose estimation.~\citep{handmesh,chen2021mobrecon} propose a novel aggregation method to collect effective 2D cues and exploit high-level semantic relations for root-relative mesh recovery. Kevin Lin proposed Graphormer~\citep{graphformer}, combining Transformer and GCN to simulate the global interaction between joints and mesh vertices.
\paragraph{Mesh-image alignment.} In the field of 2D image processing, most deep learning methods employ a "fully convolution" network framework that maintains spatial alignment between images and outputs~\citep{Pointrend,long2015fully,tompson2014joint}. Several research methods also consider alignment relationships in the 3D domain. For example, PIFu~\citep{pifu} proposed an implicit representation that locally aligns the pixels of a 2D image with the global context of their corresponding 3D objects. PyMAF~\citep{pymaf} introduced a mesh alignment feedback loop, where evidence of mesh alignment is used to correct parameters for better-aligned reconstruction results. The alignment can take advantage of more informative features that are sensitive to position to predict mesh.
\par Existing mesh recovery works~\citep{towards,li2022interacting} face the shortcomings of complex network structure when mesh-images alignment. Furthermore, the initial input of the 3D decoder is a high-resolution mesh, which makes network optimization difficult. This is critical for practical applications. To address these issues, we propose a compact network framework to map 2d image pixel features to 3d mesh vertex locations. We apply a multi-scale structure to the 2D feature encoder and 3D 
mesh decoder respectively, to achieve coarse-to-fine pixel alignment at corresponding resolutions. Using multi-scale pixel-aligned features can achieve better mesh-image alignment than previous methods.

\section{Methodology}
\par Given a monocular RGB image $I$, our goal is to predict the 3D positions of all the $N$ vertices $\vertices = \{\vertex_i\}^N_{i=1}$ of the predefined hand mesh $\mesh$. The overall architecture of our network, as shown in \Figref{fig_pipeline}, has two major components: a 2D feature extractor, as well as a 3D mesh decoder that consists of feature mapping modules and mesh-conv layers. The 2D feature extractor is an hourglass that encodes the image content into features at $S$ levels of scale. Respectively the 3D mesh decoder also recovers the vertices in a coarse-to-fine manner in $\allstages$ different scales. By design the mesh decoder at level $\stage \in \allstages$ leverages the 2D feature map at level $\stage$.  In the following sections, we will describe the architecture of the 2D feature extractor in \ref{2d_feature_extractor}, the pixel-aligned feature mapping module in \ref{Pixel-aligned_Feature_Mapping_Module}, the mesh decoder in \ref{mesh_decoder}, as well as training details in \ref{training}.
%, where %$h$ and $w$ are the height and width of the image, and $N$ is the number of vertices.

%Smaller $i$ indicates the low large resolution. These feature maps are then combined via several 1D convolution modules to form feature maps $Q _i$ for mesh-conv layers. In the feature mapping modules, mesh vertices $M_i$ are associated with image pixels $x_i$ to get mesh-aligned features $G_i$ by 3D-to-2D projection. In the 3D mesh decoder, GCN blocks aggregate the mapped local image features and the features from the previous decoder layer. Each decoder layer also estimates hand meshes of difference scales $M_i$. Next, we present three parts in detail.

% regresses mesh residuals $\delta M_i$ in a side branch. The new mesh can be represented as $M_{i+1}=Upsample(M_i+\delta M_i)$. 

% \par Given a monocular RGB image $I \in \mathbb{R}^{h*w*3}$ , our goal is to predict hand mesh $M_1 \in \mathbb{R}^{N_v*3}$, where $N_v$ is the number of vertices of the mesh. The architecture of our network is shown in Figure \ref{fig1}. It includes three parts. We first encode the input image along with other auxiliary predictions to get features $F_i$ and $H_i$. These are passed to several 1D convolution modules to concatenate as $Q _i$. In mapping module, mesh vertices $M_i$ are associated with pixels $x_i$ to get mesh-aligned features $G_i$. The GCN blocks aggregates the aligned and previous-stages features, then regress offset mesh $\delta M_i$. The new mesh can be represented as $M_{i+1}=Upsample(M_i+\delta M_i)$. Next, we present three parts in detail.

\subsection{2D Feature Extractor}
% \seclabel{2d_feature_extractor}
\label{2d_feature_extractor}
\par In order to extract 2D features at different scales/receptive fields, we adopt a simple hourglass model with skip connections as the feature extractor. Previous works~\citep{metro,weakly,cmr} extract a global image feature vector and feed it to the decoder. Since all the vertices share the same global feature, only the geometric features relating to the overall deformation of meshes are aggregated and updated for each vertex. To enable mapping local image features to the vertices, 
we combine the downsampled feature $\downsampledfeature_\stage$ and upsampled feature $\upsampledfeature_\stage$ from the respective level $\stage$ of the hourglass network to have a fusion feature map $\imagefeature_\stage \in \mathbb{R}^{h_{\stage}*w_\stage*c_{\stage}}$ for the level $\stage$ of the mesh decoder. Specifically, 
% \vspace{-0.5cm}
\begin{equation}
    Q_\stage = \mathbf{Conv1D}(\bigoplus(\downsampledfeature_\stage, \upsampledfeature_\stage)),
\end{equation}
where $\bigoplus$ denotes concatenation, $\mathbf{Conv1D}$ denotes 1D convolution. $h_\stage$, $w_\stage$, and $c_\stage$ represent height, width, and the number of channels of $\imagefeature_\stage$ respectively. 
\subsection{Pixel-aligned Feature Mapping Module}
% \seclabel{Pixel-aligned_Feature_Mapping_Module}
\label{Pixel-aligned_Feature_Mapping_Module}
\par Given a 2D image feature map $\imagefeature \in \mathbb{R}^{h*w*c}$, the mapping module needs to transform it into 3D vertex features $G \in \mathbb{R}^{N*c}$ of the corresponding mesh decoder layer, where $N$ denotes the number of vertices as mentioned. For this purpose, previous methods ~\citep{weakly,cmr, metro} either simply repeat the global feature from a feature extraction network to have the vertex features, or use a fully connected layer to map the global feature vector from $\mathbb{R}^{c}$ to $\mathbb{R}^{N*c}$ (the vertex features $G$ are obtained by reshaping $F^{'}_g$). This mapping manner can not well build the relationship between the 2D and 3D domain and have difficulty in guaranteeing mesh-alignment with the input image~\citep{pymaf}.

\par
\begin{figure}[t!]
  \centering
  \vspace{-0.8cm}
  \includegraphics[width=\textwidth,trim={0 2cm 0 2cm},clip]{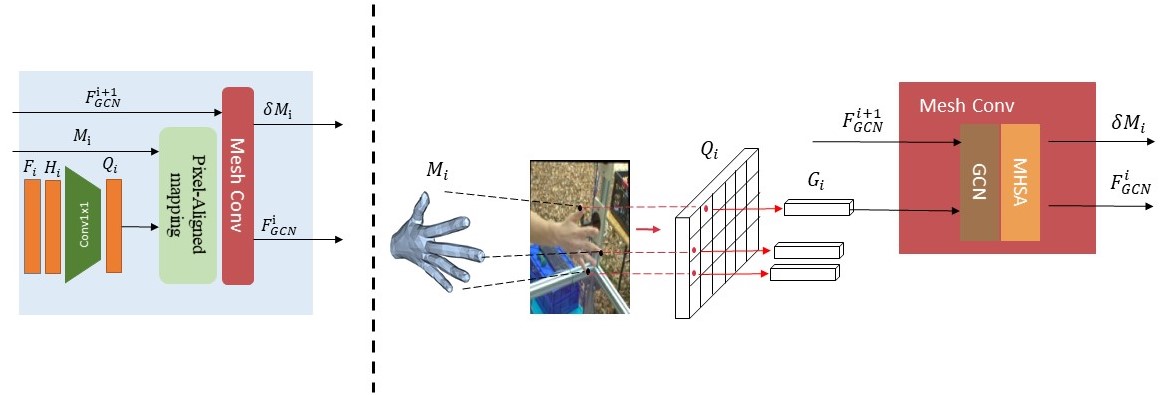}
    \vspace{-0.2cm}
  \caption{\small Left: The architecture of the mesh decoder. Right: The data flow of the mesh decoder. 3D vertices are projected to the 2D image plane to retrieve pixel-aligned features. The GCN blocks and the attention module form a mesh-conv layer to predict offset mesh.}
  \label{mesh_decoder_fig}
    \vspace{-0.2cm}
\end{figure}

 Inspired by ~\citep{pixel2mesh,pifu}, we utilize a pixel-aligned feature mapping module to transform the feature map $\imagefeature_\stage$ to 3D vertex features $G_\stage \in \mathbb{R}^{N_\stage*c_\stage}$. Each predicted vertex $\vertex \in \vertices_\stage$ is projected to pixel $\pixel$ in image space, as illustrated in \Figref{mesh_decoder_fig}. 
% Note that, we use standard template $M_4$ as a start. 
After that, we sample the feature map $\imagefeature_\stage$ to extract pixel-aligned vertex features $\meshfeature_\stage$ using the following equation:
\begin{equation}
\begin{split}
    \meshfeature_\stage = f(\imagefeature_\stage, \pi (\mesh_\stage)),
\end{split}
\end{equation}
where $\pi (.)$ denotes 2D projection, and $f(.)$ is a sampling function. 
\par Similar to ~\citep{pixel2mesh,pymaf} we use bilinear interpolation around each projected vertices on the feature maps to extract pixel-aligned feature vectors. The 2D feature maps $\{\imagefeature_\stage\}$ are multi-scale (7x7->14x14->28x28->56x56). Lower-resolution image features have a larger receptive field, hence more global information, while higher-resolution feature maps contain more local information. The pyramid feature maps and pixel-aligned feature mapping modules can provide richer content for vertices.

\subsection{Mesh Decoder}
% \seclabel{mesh_decoder}
\label{mesh_decoder}
In the 2D feature extractor, feature maps of different scales are kept and in the mapping module, the connections of the features of pixels of these feature maps and features for the 3D vertices in the corresponding decoding layers are established. In this section, we aim to design a mesh decoder structure that makes full use of the connections that are made possible via the non-parametric mesh representation. To be specific, we aim to 1) design a mechanism to leverage the multi-scale image features in the hierarchical decoding architecture; 2) aggregate both local features from the feature extractor and the features extracted in the previous mesh decoder layers.

For the first aim, our decoder reconstructs hand meshes in a coarse-to-fine manner along with the hierarchical layers, shown in \Figref{fig_pipeline}. Our blocks skip connects the different vertex feature $G_i$ from the mapping module. In each decoding layer, based on the reconstructed mesh, image features of the corresponding scale are sampled by the feature mapping module and then concatenated to the vertices features. Rather than directly regressing the 3D coordinates, we predict the offset mesh $ \delta{M_i}$ of last-block-estimated ${M_i}$.
The coarse-to-fine mesh topologies for each block are acquired by mesh simplification~\citep{spiralnet++}. The number of vertices of different meshes is 98, 195, 389, and 778. The $M_1$ has the same mesh topology as MANO~\citep{mano}. The meshes at different scales are predicted by several additional mesh decoding headers. We use the strategy based on edge contraction~\citep{garland1997surface} to do the pooling and unpooling operation. The input features are multiplied with a pre-computed transform matrix to obtain the output features.

We achieve the second aim by deploying operations that can work on vertices of meshes, \ie graph convolution operating on the manifold of the meshes and the attention module. In the ablation studies, we demonstrate that local features can significantly boost the performance of both operations compared with global features. 
Though in our approach, the feature mapping and the hierarchical decoding are only applied to decoders with graph convolutions and the attention modules, they can be injected into other 3D mesh decoders. In the experiment, we show that our components are also very effective for the transformer-based method~\citep{metro}.
\paragraph{Graph Convolution} For the graph convolution operators, spectral convolutions and spatial convolutions have been used in the mesh reconstruction. We use a spiral patch operator~\citep{spiralnet++} to process vertex features in the spatial domain as it demonstrates superior performance in recent work. 
The spiral operator collects vertex neighbors of the center $\vertex$. The graph convolution produces the offset mesh $\delta \mesh_\stage$ and feature $F_{GCN}^\stage$ for each vertex. The spiral convolution of input $F_{GCN}^\stage(\vertex)$ can be defined as:
\begin{equation}
\begin{split}
    (F_{GCN}^\stage *\gcnkernel)_\vertex=\sum_{\seq=1}^{\seqlength}\gcnkernel_\seq F_{GCN}^\stage(\seqorder(\vertex))
\end{split}
\end{equation}
where $\gcnkernel_\seq$ is the convolution kernel, and $\seqorder(\vertex)$ denotes the pre-computed vertices ordered sequence.
\paragraph{Self-Attention} While GCN is useful for capturing fine-grained neighborhood information, it is less efficient at extracting long-range dependencies. We inject the multi-head self-attention module~\citep{mhsa} into GCN blocks to address this challenge.
\par Given the vertex features $F_{GCN}^\stage$ generated by GCN, we strengthen the global interactions and get a new feature $F_{MHSA}^\stage$ for each vertex with the help of MHSA:
\begin{equation}
\begin{split}
    F_{MHSA}^i=MHSA(F_{GCN}^\stage),
\end{split}
\end{equation}
The GCN blocks and the attention module form a mesh-conv layer as shown in \Figref{mesh_decoder_fig}.

\subsection{Training}
% \seclabel{training}
\label{training}
\par We denote a dataset as a set of tuples $\{(I, \poseheatmap, \silhouetteheatmap, \mesh)\}$, where $I$ is the input image and $\mesh$ is the hand mesh; $\poseheatmap$ and $\silhouetteheatmap$ are the heatmaps of 2D pose and silhouette, respectively. Following ~\citep{meshnet}, We apply the Gaussian distribution to construct 2D pose heatmaps.

\paragraph{3D Supervision} One concern for non-parametric methods is the lack of kinematic and shape constraints. Adding 3D shape supervision can mitigate that. To this end, we adopt an L1 loss norm $L_{mesh}$ to supervise the 3D coordinates of vertices in the coarse-to-fine manner. Besides, we use the normal loss $L_{norm}$ and edge length loss $L_{edge}$ in ~\citep{pixel2mesh} for smoother reconstruction meshes in the last mesh decoding layer. So the loss for a sample in the dataset is defined as:
\begin{equation}
\begin{split}
    L_{mesh} &= \sum _{\stage=1} ^{\allstages}\lambda_m^s*||\hat{\mesh_\stage}-\mesh_\stage||_1 \\ 
    L_{edge} &= \sum _{\face \in \allfaces}\sum _{\edge \in \face} || |\hat{\edge}|-|\edge|| ||_1\\
    L_{norm} &= \sum _{\face \in \allfaces}\sum _{\edge \in \face} || \hat{\edge} \cdot \facenormal ||_1\\
\end{split}
\end{equation}
where $\face$ is a triangle face from all the faces $\allfaces$ of $\mesh$, $e$ denotes an edge of $\face$, and $\facenormal$ the normal vector of $\face$ computed from the ground truth. To ensure the performance of the reconstruction of the finest mesh, we set a weight $\lambda_m^s$ for different scale mesh recontruction loss.

\paragraph{2D Auxiliary Supervision} For the 2D feature extractor, we add auxiliary supervision for better feature extraction. We apply binary-cross-entropy (BCE) loss to formulate both the silhouette loss $L_{sil}$ and 2D pose loss $L_{2Dpose}$ as follows:
\begin{equation}
\begin{split}
    L_{sil} &= -(\sum _j(\silhouetteheatmappixel_j \log(\hat{\silhouetteheatmappixel_j})+(1-\silhouetteheatmappixel_j) \log(1-\hat{\silhouetteheatmappixel_j}))), \silhouetteheatmappixel \in \silhouetteheatmap\\ 
    L_{2Dpose} &= -(\sum _j(\poseheatmappixel_j \log(\hat{\poseheatmappixel_j})+(1-\poseheatmappixel_j) \log(1-\hat{\poseheatmappixel_j}))), \poseheatmappixel \in \poseheatmap,
\end{split}
\end{equation}
where $\silhouetteheatmappixel_j$ and $\poseheatmappixel_j$ denotes the $j$-th pixel value of the silhouette and pose heatmaps respectively, and~ $\hat{}$ denotes the prediction.

The overall loss is a weighted sum of all losses, $L_{total}=L_{mesh}+L_{edge}+\lambda _n* L_{norm}+\lambda _s*L_{sil}+ \lambda _p*L_{2Dpose}$, where $\lambda _n=0.1$, $\lambda _p=10$, $\lambda _s=2.5$ are set empirically.

\section{Experiment}
\subsection{Experimental Setups}
\label{sec:4.1}
\par \textbf{FreiHAND}~\citep{freihand} is the most widely-used benchmark for hand mesh reconstruction. It contains 130K images for training with 3D and 2D annotations, and the test set has 4K images. As the annotations of the test set are not available, we submit our results to their provided online server for evaluation.

\paragraph{Training Procedure.} Our network is based on the HRNet-W64~\citep{hrnet} and ResNet-50~\citep{resnet} backbone pre-trained on ImageNet~\citep{imagenet}. We trained our model in an end-to-end manner. 2D keypoints and masks are used to train the image feature extractors as auxiliary supervision. We resize the input image to 224$\times$224.
Following previous work, data augmentations including rotation, translation, color jitter, etc are applied during training. Our method is implemented in Pytorch and trained on Nvidia RTX 3090 GPU with a batch size of 64 for 50 epochs. We optimize the network by Adam with a learning rate of 1e-4 and set a decay rate of 0.1 at 35 epochs. 

\paragraph{Evaluation metrics.} We report the results of our approach using several evaluation metrics. PA-MPJPE: It first performs the rigid alignment between the prediction and ground-truth using Procrustes Analysis~\citep{PA}, then calculates the mean-per-joint-position-error. PA-MPVPE: similar to PA-MPJPE, but it measures the difference between the vertices predicted and the ground truth. The unit for the PA-MPJPE/PA-MPVPE metrics is millimeter (mm). F-scores: It measures the harmonic mean between recall and precision between two meshes. We report the F@5mm and F@15mm as existing works. 
\begin{table}[!t]
\vspace{-1.2cm}
  \caption{\small Analysis of the self-attention and mapping module}
  \label{table2}
  \centering
  \begin{tabular}{lllll}
    \\
    \toprule
    \cmidrule(r){1-5}
    Feat.     &Mapping Module &Self-attention &PA-MPJPE$\downarrow$ &PA-MPVPE$\downarrow$\\
    \midrule
        Global   &Repeat &\XSolidBrush &10.1 &10.3\\
    Global   &MLP &\XSolidBrush &9.1 &9.2\\
    Feature maps(w/o 2D) &Pixel-aligned &\XSolidBrush   &8.3 &8.4\\
    Feature maps(w/ 2D) &Pixel-aligned &\XSolidBrush   &7.8 &7.9\\
    Global   &MLP &\CheckmarkBold       &8.4 &8.6\\
    Feature maps &Pixel-aligned &\CheckmarkBold   &7.5 &7.7\\
    \bottomrule
\vspace{-0.8cm}
  \end{tabular}
  \end{table}
% \begin{table}[!t]
% % \vspace{-1.5cm}
%   \caption{\small Analysis of the self-attention and mapping module}
%   \label{table2}
%   \begin{center}
%   \begin{tabular}{lllll}
%     \multicolumn{1}{c}{Feat.} &\multicolumn{1}{c}{Mapping Module} &\multicolumn{1}{c}{Self-attention} &\multicolumn{1}{c}{PA-MPJPE$\downarrow$} &\multicolumn{1}{c}{PA-MPVPE$\downarrow$}
%     \\ \hline \\
%     Global   &Repeat &\XSolidBrush &10.1 &10.3\\
%     Global   &MLP &\XSolidBrush &9.1 &9.2\\
%     Feature maps(w/o 2D) &Pixel-aligned &\XSolidBrush   &8.3 &8.4\\
%     Feature maps(w/ 2D) &Pixel-aligned &\XSolidBrush   &7.8 &7.9\\
%     Global   &MLP &\CheckmarkBold       &8.4 &8.6\\
%     Feature maps &Pixel-aligned &\CheckmarkBold   &7.5 &7.7\\
%   \end{tabular}
%   \end{center}
%   \end{table}

\begin{minipage}{\textwidth}
\begin{minipage}[t]{0.45\textwidth}
\makeatletter\def\@captype{table}
% \begin{table}[!t]
  \caption{\small Analysis of adding skip connections in different decoder layers.}
  \label{removefeature}
  \centering
  \begin{tabular}{llllll}
    \\
    \toprule
    \cmidrule(r){1-6}
    $M_4$   &$M_3$ &$M_2$ &$M_1$ &PJ$\downarrow$ &PV$\downarrow$\\
    \midrule
    \CheckmarkBold &\XSolidBrush &\XSolidBrush &\XSolidBrush &8.25 &8.31\\
    \CheckmarkBold &\CheckmarkBold &\XSolidBrush &\XSolidBrush &8.06 &8.13\\
    \CheckmarkBold &\CheckmarkBold &\CheckmarkBold &\XSolidBrush &7.93 &8.02\\
    \CheckmarkBold &\CheckmarkBold &\CheckmarkBold &\CheckmarkBold  &7.83 &7.93\\
    \bottomrule
  \end{tabular}
% \end{table}
\end{minipage}
\hspace{2em}
\begin{minipage}[t]{0.45\textwidth}
\makeatletter\def\@captype{table}
% \begin{table}[!t]
  \caption{\small Effectiveness of coarse-to-fine for pixel-aligned mapping module}
  \label{compare_other_pixel}
  \centering
  \begin{tabular}{lll}
    \\
    \toprule
    \cmidrule(r){1-3}
    Methods &PJ$\downarrow$ &PV$\downarrow$\\
    \midrule
    refine(A time) &8.31 &8.42\\
    refine(Three times)   &8.12 &8.22\\
    \midrule
    \textbf{Ours} &7.83 &7.93\\
    \bottomrule
  \end{tabular}
% \end{table}
\end{minipage}
\end{minipage}
% \begin{table}[!t]
%   \caption{\small Analysis of adding skip connections in different decoder layers.}
%   \label{removefeature}
%   \begin{center}
%   \begin{tabular}{llllll}
%     \multicolumn{1}{c}{\bf $M_4$} &\multicolumn{1}{c}{\bf $M_3$} &\multicolumn{1}{c}{\bf $M_2$} &\multicolumn{1}{c}{\bf $M_1$} &\multicolumn{1}{c}{\bf PJ$\downarrow$} &\multicolumn{1}{c}{\bf PV$\downarrow$}
%     \\ \hline \\
%     \CheckmarkBold &\XSolidBrush &\XSolidBrush &\XSolidBrush &8.25 &8.31\\
%     \CheckmarkBold &\CheckmarkBold &\XSolidBrush &\XSolidBrush &8.06 &8.13\\
%     \CheckmarkBold &\CheckmarkBold &\CheckmarkBold &\XSolidBrush &7.93 &8.02\\
%     \CheckmarkBold &\CheckmarkBold &\CheckmarkBold &\CheckmarkBold  &7.83 &7.93\\
%   \end{tabular}
% \end{center}
% \end{table}

% \begin{table}[!t]
%   \caption{\small Effectiveness of coarse-to-fine for pixel-aligned mapping module}
%   \label{compare_other_pixel}
%   \begin{center}
%   \begin{tabular}{lll}
%     \multicolumn{1}{c}{\bf Methods} &\multicolumn{1}{c}{\bf PJ$\downarrow$} &\multicolumn{1}{c}{\bf PV$\downarrow$}
%         \\ \hline \\
%     refine(A time) &8.31 &8.42\\
%     refine(Three times)   &8.12 &8.22\\
%     Ours &7.83 &7.93\\
%   \end{tabular}
% \end{center}
% \end{table}

\subsection{Ablation Studies}
\label{ablation}
\par We conduct ablation studies under various settings on FreiHAND~\citep{freihand} to investigate the key components of our model. We use ResNet-18 as the backbone and report the results using PA-MPJPE and PA-MPVPE. Table~\ref{table2} shows all the comparisons. Our final model (last row) improves the baseline with the global feature (1st row) by 26$\%$.
\paragraph{Effectiveness of Mapping Module for Graph Convolution} To establish the relationships between the 2D pixels and 3D vertices, We utilize the pixel-aligned feature mapping module. To verify the efficacy, we construct baselines based on GCN and compare their reconstruction accuracy in Table~\ref{table2} with the variant of our method. Similar to ~\citep{weakly}, we implement a baseline that directly repeats the global feature and concatenates it to the mesh vertices (1st row) and a baseline that maps the global feature to the vertex features by MLP layers(2nd row). For fair comparisons, we construct our variant with the pixel-aligned module which has the same architecture for the mesh decoder with these two baselines and do not add the 2D supervision (3rd row). Table~\ref{table2} shows that the pixel-aligned module improves the reconstruction performance by a large margin, about 18$\%$ and 8$\%$.

% In our approach, the auxiliary predictions (2D pose and silhouette) can provide better context for convolution on 3D vertices.
\paragraph{Effectiveness of Attention and Mapping Module for Attention} Adding attention to graph convolution can complement the long-range information aggregation and hence improve the network capacity. To verify this, we compare a network using the graph convolution only (2nd row) and a network with an attention module in between graph convolution layers (5th row) in Table~\ref{table2}. The results show that the attention mechanism can improve the mesh reconstruction with only global features. Based on the same graph attention architecture, adding the mapping module can further improve the performance by 0.9mm in PA-MPJPE. 

\paragraph{Effectiveness of 2D Auxiliary Supervision} We analyze how feature maps affect. We implement two models with the same architecture: one has 2D supervision from silhouette and 2D pose (4th row) and one without (3rd row). It shows that the feature maps with supervision are helpful for improving performance. As 3D mesh annotation is expensive to acquire while 2D pose can be relatively cheap to label, we expect our method can benefit more from extra 2D auxiliary supervision.

\paragraph{Adding Skip Connections in Different Decoder Layers} To analyze the finer local features in the mesh reconstruction, variants of our method are constructed by gradually removing the skip connections in the last decoding layer and the results of these variants are shown in Table~\ref{removefeature}. Note that, all approaches do not add the self-attention module. We observe a gradual reduction in the errors when multi-scale mesh-alignment evidences are removed from decoder layers. It proves that our skip connection design improves accuracy by leveraging multi-scale information.

\paragraph{Effectiveness of Coarse-to-fine for Pixel-aligned Mapping Module} We reproduced \citep{towards} pipeline with our mesh decoder architecture to compare with another pixel-aligned-based method. Rather than regress hand mesh vertices in a coarse-to-fine manner like ours, they use the full mesh to refine. In the first row, we follow their pipeline to refine the rough mesh one time, and for a more intuitive comparison with our method, we refine the mesh three times. Table~\ref{compare_other_pixel} shows our coarse-to-fine manner can achieve better results for both two cases.

% \begin{table}[!t]
%   \caption{\small The generality of mapping module and skip connections}
%   \label{non-para results}
%   \begin{center}
%   \begin{tabular}{lll}
%     \multicolumn{1}{c}{Methods} &\multicolumn{1}{c}{PA-MPJPE$\downarrow$} &\multicolumn{1}{c}{PA-MPVPE$\downarrow$}
%         \\ \hline \\
%     METRO\citep{metro} &6.8 &6.7\\
%     METRO with our designs   &6.4 &6.5\\
%     Graphormer\citep{graphformer} &6.0 &5.9 \\
%     Graphormer with our designs & &\\
%   \end{tabular}
% \end{center}
% \end{table}
\begin{table}[!t]
 \vspace{-1.2cm}
  \caption{\small Performance comparison with the state-of-the-art methods on the FreiHAND dataset.$\downarrow$ means the lower the better, $\uparrow$ means the higher the better. The unit of PA-MPJPE/PA-MPVPE is mm.}
  \label{table1}
  \begin{center}
  \begin{tabular}{lllll}
  \\
   \toprule
    \cmidrule(r){1-5}
    Methods &PA-MPJPE$\downarrow$ &PA-MPVPE$\downarrow$ &F@5mm$\uparrow$ &F@15mm$\uparrow$ \\
    \midrule
    \citep{hasson}  &13.3 &13.3 &0.429 &0.907\\
    \citep{freihand}  &10.9 &11.0 &0.516 &0.934\\
    \citep{weakly}  &8.4 &8.6 &0.614 &0.966\\
    Pose2Mesh\citep{pose2mesh}  &7.7 &7.8 &0.674 &0.969\\
    I2LMeshNet\citep{meshnet}  &7.4 &7.6 &0.681 &0.973\\
    \citep{towards} &7.1 &7.1 &0.706 &0.977 \\
    \citep{handmesh} &6.9 &7.0 &0.715 &0.977 \\
    METRO\citep{metro} &6.8 &6.7 &0.717 &0.981\\
    Graphormer\citep{graphformer} &6.0 &5.9 &0.764 &0.986 \\
    \midrule
    \textbf{Ours with METRO}        &\textbf{6.1} &\textbf{6.2} &\textbf{0.757} &\textbf{0.984}\\
    \textbf{Ours with GCN}        &\textbf{5.9} &\textbf{6.0} &\textbf{0.766} &\textbf{0.985}\\
    \bottomrule
    \vspace{-0.8cm}
    \end{tabular}
    \end{center}
\end{table}
\paragraph{The Generality of Mapping Module and Skip Connections} We verify that our pixel-aligned mapping module and skip connections are not only effective for our approach but also can improve the performance 
of other non-parametric models. Based on METRO~\citep{metro} which are transformer-based methods, we re-implement their networks with our design. We provide the details of network architecture in the appendix due to the space limitation. As Table~\ref{table1} shows, pixel-aligned mapping module and skip connections can improve the METRO by a large margin.
% \jsj{add the result of adding pixel-aligned for transformer, explain it is also very useful. The result of Graphormer to be added}
% \begin{figure}[!htbp]
%   \centering
%   \includegraphics[width=\textwidth,trim={0 0cm 0 0cm},clip]{figs/acc.jpg}
%   \caption{\small Left: comparison of the baseline with global features and our method; Right: Example meshes from $M_4$ to $M_1$.}
%   \label{fig_acc}
% \end{figure}
\subsection{Comparisons with State-of-the-Art}
\label{sec:4.3}
% \jsj{Remove the result of model size and GPU-effiency\\}
We follow the former works~\citep{freihand,hasson,weakly,pose2mesh,meshnet,metro,graphformer} to quantitatively compare our method with state-of-the-art methods on the FreiHAND eval set. For GCN, we re-design our 2D feature extractor as CMR~\citep{handmesh} for further performance improvement. For Transformer, we design the architecture as introduced in \ref{ablation}. As shown in Table \ref{table1}, All of our approaches achieve state-of-the-art for all the metrics. It demonstrates that our designs can be effective for both non-parametric models.

\begin{figure}[!t]
  \centering
  \vspace{-1cm}
  \includegraphics[width=\textwidth,trim={0 1cm 0 0.5cm},clip]{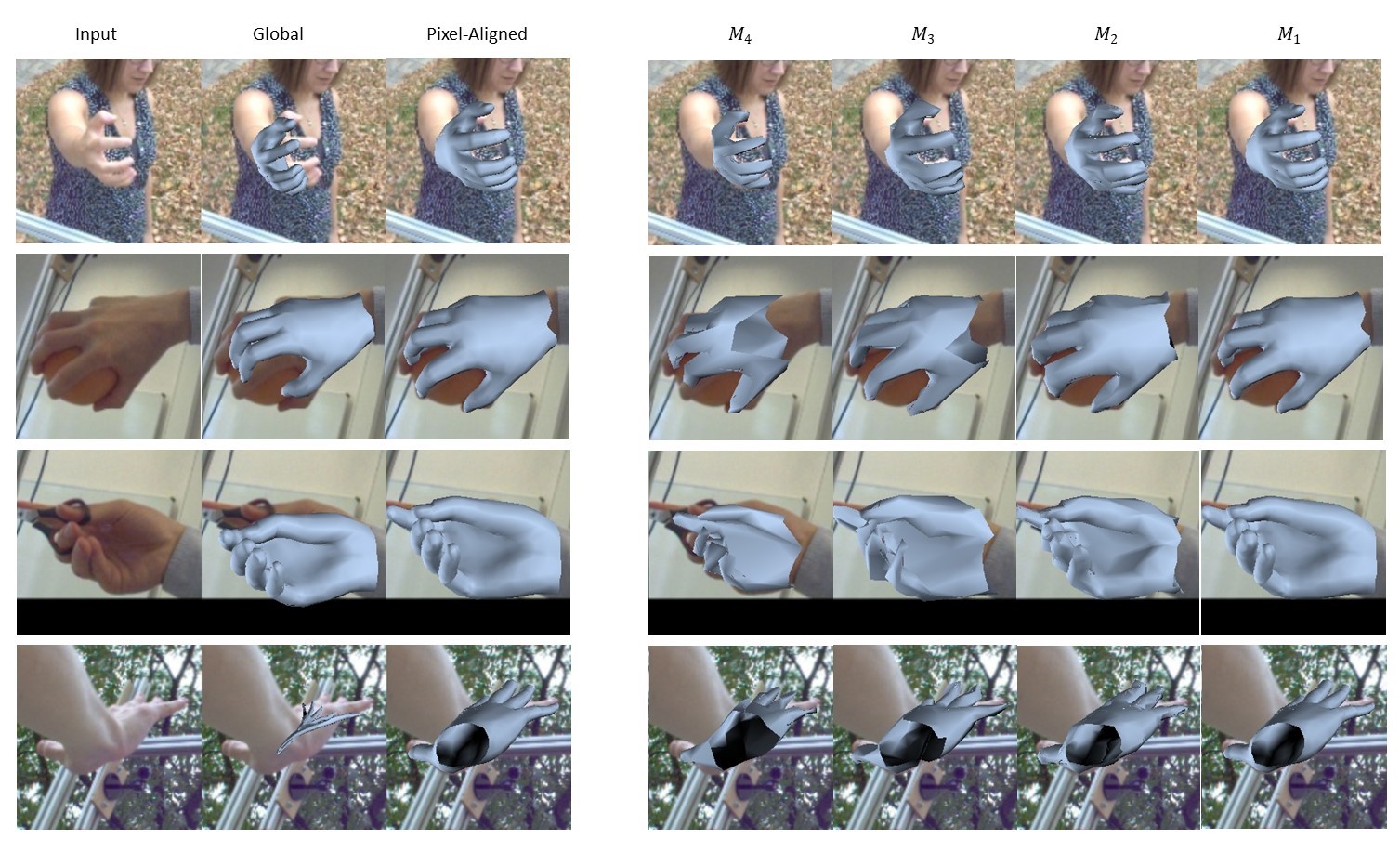}
  \vspace{-0cm}
  \caption{\small Left: comparison of the baseline with global features and our method; Right: Example meshes from $M_4$ to $M_1$.}
  \label{fig_examples}
        \vspace{-0.2cm}
\end{figure}

\paragraph{Qualitative Results} \Figref{fig_examples} shows reconstructed meshes from several testing examples of FreiHAND. On the left side, our method aligns the meshes better to the input images than the baseline with global features only. On the right side, decoded meshes in different scales are shown. Notice that the meshes from the last layers adjust both the global locations to align the mesh to the inputs (\eg the meshes in the 3rd row) and the geometry of the meshes (\eg the thickness and smoothness of the meshes). 
\par
\begin{figure}[!t]
  \centering
  \includegraphics[width=0.7\textwidth,trim={0 1cm 0 0cm},clip]{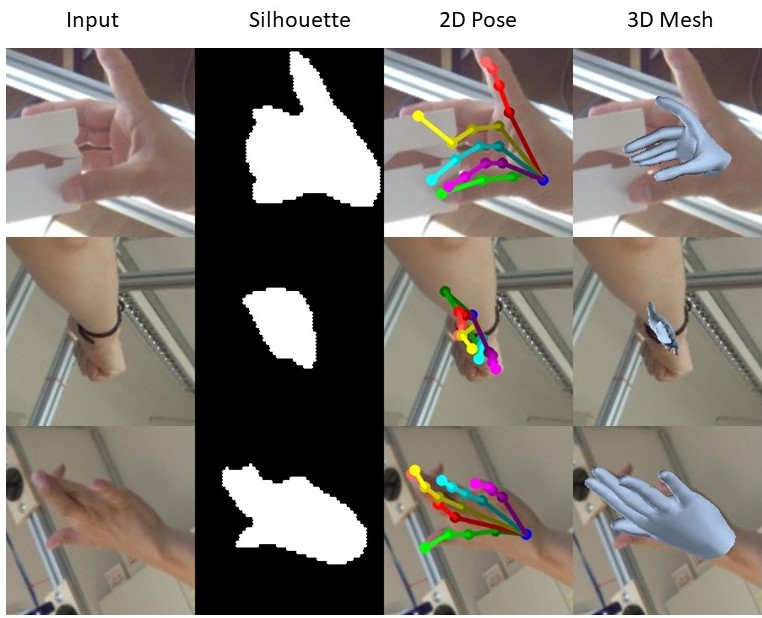}
  \caption{\small Failure cases of our method.}
  \label{fig_failure}
  \vspace{-0.5cm}
\end{figure}

 \Figref{fig_failure} shows three typical failure cases of our method. In the first row, when the hand is severely occluded and the hands are not bounded by the crop size,  some parts of the hand out of the image, our method fails to recover a correct hand mesh. In the second row, we observe when only a small portion of the hand is visible, our method predicts the wrong 2D pose and silhouette as well as the hand mesh. Referring to the last row, although the overall shape seems to be reasonable, it is difficult to obtain an accurate 3D mesh due to the heavy self-occlusion. The self-occlusion is one of the biggest challenges for 3D hand mesh reconstruction or pose estimation.
% \vspace{-0.2cm}

\section{Conclusions}
% \par We present a new pipeline to reconstruct a hand from a single RGB image. Specifically, we introduce a novel pixel-aligned feature mapping module that can help align the mesh to the image, together with adding the self-attention module to improve vertices interactions. Comprehensive experiments show our method achieves the state-of-the-art on the FreiHAND dataset and verifies the effectiveness of our mapping module.
\par We present a new pipeline to reconstruct a hand from a single RGB image. Specifically, we introduce a simple and compact architecture that can help align the mesh to the image, together with adding the self-attention module to improve vertices interactions. Comprehensive experiments show our method achieves the state-of-the-art on the FreiHAND dataset and verifies the effectiveness of our proposed key components. We further demonstrated that our design can also improve the performance of the transformer-based method. It shows that our proposed components have great generality for non-parametric models.

\newpage
\bibliography{iclr2023_conference}
\bibliographystyle{iclr2023_conference}
% \newpage
% \appendix
% \input{text/appendix.tex}

\end{document}